\documentclass[conference]{IEEEtran}
\IEEEoverridecommandlockouts
\usepackage{cite}
\usepackage{amsmath,amssymb,amsfonts}
\usepackage{graphicx}
\usepackage{graphics,subfig}
\usepackage{textcomp}
\usepackage{multirow}
\usepackage{xcolor}
\usepackage{amsthm}
\usepackage{comment}
\usepackage{url}
\usepackage{caption}
\usepackage[ruled,vlined]{algorithm2e}
\usepackage{framed,enumitem}
\usepackage{verbatim}
\usepackage{hyperref}
\usepackage{multirow}
\usepackage{float}
\hypersetup{
    colorlinks=true,
    linkcolor=blue,
    filecolor=magenta,      
    urlcolor=blue,
}
\usepackage{array}
\newcolumntype{M}[1]{>{\centering\let\newline\\\arraybackslash\hspace{0pt}}m{#1}}
\usepackage{algpseudocode}

\def\BibTeX{{\rm B\kern-.05em{\sc i\kern-.025em b}\kern-.08em
    T\kern-.1667em\lower.7ex\hbox{E}\kern-.125emX}}
    
\DeclareMathOperator*{\argmax}{arg\,max}

\newcommand{\E}{\mathbb{E}}

\newcommand{\R}{\mathbb{R}}

\newcommand{\calA}{\mathcal{A}}
\newcommand{\calB}{\mathcal{B}}
\newcommand{\calC}{\mathcal{C}}

\newcommand{\calS}{\mathcal{S}}
\newcommand{\calX}{\mathcal{X}}

\newcommand{\calG}{\mathcal{G}}
\newcommand{\calR}{\mathcal{R}}

\newcommand{\calN}{\mathcal{N}}
\newcommand{\calE}{\mathcal{E}}
\newcommand{\calT}{\mathcal{T}}

\newcommand{\calP}{\mathcal{P}}

\newcommand{\calF}{\mathcal{F}}

\newcommand{\calU}{\mathcal{U}}

\theoremstyle{definition}
\newtheorem{definition}{Definition}

\theoremstyle{remark}
\newtheorem{remark}{Remark}

\begin{document}
\title{Automated Adversary Emulation for Cyber-Physical Systems via Reinforcement Learning}
\author{
\IEEEauthorblockN{
Arnab Bhattacharya\IEEEauthorrefmark{1}, 
Thiagarajan Ramachandran\IEEEauthorrefmark{1}, 
Sandeep Banik\IEEEauthorrefmark{3}, 
Chase P. Dowling\IEEEauthorrefmark{2}, 
Shaunak D. Bopardikar\IEEEauthorrefmark{3}}
\IEEEauthorblockA{
\IEEEauthorrefmark{1}\textit{Optimization and Control Group}, \IEEEauthorrefmark{2}\textit{Information Modeling and Analytics Group} \\ 
\textit{Pacific Northwest National Laboratory}, Richland, WA, USA}
\IEEEauthorblockA{\IEEEauthorrefmark{3}\textit{Electrical and Computer Engineering}, 
\textit{Michigan State University}, East Lansing, MI, USA}
}
\maketitle
\begin{abstract}
Adversary emulation is an offensive exercise that provides a comprehensive assessment of a system's resilience against cyber attacks. However, adversary emulation is typically a manual process, making it costly and hard to deploy in cyber-physical systems (CPS) with complex dynamics, vulnerabilities, and operational uncertainties. In this paper, we develop an automated, domain-aware approach to adversary emulation for CPS. We formulate a Markov Decision Process (MDP) model to determine an optimal attack sequence over a hybrid attack graph with cyber (discrete) and physical (continuous) components and related physical dynamics. We apply model-based and model-free reinforcement learning (RL) methods to solve the discrete-continuous MDP in a tractable fashion. As a baseline, we also develop a greedy attack algorithm and compare it with the RL procedures. We summarize our findings through a numerical study on sensor deception attacks in buildings to compare the performance and solution quality of the proposed algorithms.
\end{abstract}
\section{Introduction}\label{sec:intro}

With increasing sophistication of today's cyber attacks, there is a critical need for offensive testing to assess the resilience of cyber systems. Such offensive exercises come in different flavors. Pre-compromise tests, such as penetration testing, involve probing a system to identify vulnerabilities under controlled rules of engagement. By contrast, post-compromise exercises, such as adversary emulation, require a team of cybersecurity experts - called a red team - to emulate end-to-end attacks following a set of realistic tactics, techniques, and procedures (TTPs). Compared to penetration testing, adversary emulation provides a complete security assessment to identify, contain and mitigate cyber threats. Moreover, adversary emulation generates attack scenarios that can be used to verify and improve post-compromise resilience. However, current adversary emulation requires a highly-skilled red team to manually draft the attack sequences, which can be time-consuming, costly, and personnel constrained~\cite{miller2018,Applebaum2017}.

Adversary emulation is even more challenging for cyber-physical systems (CPS), which integrate computing resources, communication protocols and physical processes. With rapid infiltration of Internet-of-Things (IoT) devices and smart sensors, red teams have to deal with an ever expanding attack surface. Unlike cyber systems, scant forensic evidence exists of post-compromise breaches in CPS, which makes it difficult to plan and execute emulation tests. Moreover, CPS operate under multiple sources of uncertainty that need to be characterized in emulation exercises. This is challenging for complex CPS where red teams have limited domain knowledge and the number of operational scenarios can be prohibitively large~\cite{Applebaum2017}. Finally, adversary emulation is risky during online operations of critical CPS (e.g. hospitals, power grid) as possible equipment damages and service disruptions can cause widespread economic loss and safety hazards~\cite{Angle2019}. The aforesaid challenges create a pressing need for automated emulation tools for CPS.

Attack graphs are traditionally used to develop emulation tools for cyber systems~\cite{Applebaum2016}. An attack graph models vulnerabilities in networked systems via  series of discrete exploits that lead to a compromised security state~\cite{Munoz-Gonzalez2015}. Efficient greedy algorithms with adversarial performance guarantees exist for attack graphs with specific topological constraints~\cite{nguyen2016targeted}. 
Moreover, game theory has been extensively used to harden cyber networks via analysis on attack graphs~\cite{etesami2019dynamic}. It is noted that hybrid attack graphs are more suitable to model the discrete (cyber) and continuous (physical) components of cross-domain CPS attacks~\cite{Ibrahim2019}. Recently, advances in adversarial machine learning, such as generative adversarial networks, have been used to design intrusion detection systems that are robust to adversarial data perturbations in CPS~\cite{Chhetri2019}. 




While optimal attack sequences can be determined for certain classes of CPS \cite{chen2017optimal}, the general problem of adversary emulation needs a principled application of the philosophy of ``thinking like an attacker". A key gap in the literature on CPS security is the assumption of an omniscient attacker, which is unrealistic for many complex CPS and may lead to extremely conservative defensive postures~\cite{etesami2019dynamic}. Moreover, it is essential to consider a cross-layer viewpoint of adversary emulation due to strong security inter-dependencies between the cyber and physical layers.
Existing studies do not adequately address the coupling between cross-layer vulnerabilities and physical dynamics for adversary emulation in CPS.

We present a new domain-aware, reinforcement learning based approach to automated adversary emulation for CPS. The key contributions of this work are summarized as follows. First, we formulate a novel Markov Decision Process (MDP) model to determine an optimal attack strategy over a hybrid attack graph with cyber (discrete) and physical (continuous) components and domain-specific dynamics. Second, we develop two competing solution procedures, based on model-based and model-free reinforcement learning (RL), to approximately solve the MDP model with a hybrid state space.  Third, we design a greedy attack algorithm that exploits the topology of hybrid attack graphs and serves as a baseline for the RL procedures. Finally, we demonstrate the performance and solution quality of the proposed algorithms on  a use-case involving sensor deception attacks on buildings. 
\section{Model Formulation}\label{sec:modelformulation}




\subsection{Hybrid Attack Graph}\label{sec:attackgraph}
A hybrid attack graph (HAG) models the security state space of a CPS, where the nodes represent security attributes (or capabilities), while the edges denote adversarial exploits (or actions). The leaf nodes describe entry-point cyber attributes, edges denote cyber exploits, and root nodes signify target physical attributes. Each attack action (cyber exploit or physical attack) has associated success probability, reward and cost values. We restrict our analysis to directed acyclic HAGs that enforce the well-known \textit{monotonicity} assumption~\cite{Miehling2015}, which states that an adversary never willingly relinquishes attributes once obtained. Next, we formally define a HAG.
\begin{definition}[{\bf HAG}]\label{def:hag}
A directed acyclic hybrid attack graph is a tuple $\calG=(\calN, \calE, \calF, \calA, \calR, \calC, \Phi)$, where
\begin{enumerate}
   \item $\calN = \{1, 2, \dots, N\}$ is the set of attribute nodes. The set of cyber and physical nodes is denoted by $\calC$ and $\calP$, respectively, such that $\calN=\calC\cup\calP$.
   \item $\calE$ is the set of edges describing unique cyber exploits. 
   \item $\calF$ is the set of functions governing the dynamics at the physical nodes, where $f_n\in\calF$ is the dynamics at $n\in\calP$.
   \item For $e\in\calE$, let  $a_e$ be the cyber exploit along edge $e$.  Define $\calA_\calE:=\cup_{e\in\calE}a_e$ to be the set of all cyber exploits. Let $\calA_n$ be the set of available actions at a physical node $n\in\calP$. Denote the act of doing nothing by $\varnothing$. Then, the set of all attack actions in $\calG$ is $\calA =\{\varnothing\}\cup\calA_\calE\cup\bigl(\cup_{n\in\calP}\calA_n\bigr)$,    where $\calA$ is assumed to be finite.

   \item $\calR=\{r_e, r_n\}_{e\in \calE, n\in\calP}$ is the set of real-valued reward functions, where $r_e^a$ (resp. $r_n^a$) is the reward for selecting action $a$ along edge $e$ (resp. at physical node $n$). The corresponding set of cost functions is $\calC=\{c_e, c_n\}_{e\in\calE, n\in\calP}$.
    \item $\Phi:\calA\times\calT\rightarrow[0,1]$ is a probability mass function that maps an action $a$ at time $t$ to its success probability $\Phi_t^a$.
\end{enumerate}
\end{definition}
\begin{remark}[{\bf Relationship with pre-conditioning}]
Definition \ref{def:hag} ignores the notion of the so-called  pre-conditions~\cite{saha2016identifying}, where access to nodes with an AND pre-condition require all origin nodes to be compromised, while an OR pre-condition requires at least one is compromised. We assume that each node in a HAG has an OR pre-condition only; this can be easily relaxed at the expense of notational complexity.
\end{remark}
\begin{remark}[{\bf Attack success probabilities}]
It is assumed that the attack-success probabilities are governed by a time-varying and non-adaptive defender policy, which is independent of the attack policy (see~\cite{Hota2018} for a similar setup).  
\end{remark}
\begin{figure}[htb!]
    \vspace{-3mm}
    \centering
    \includegraphics[scale=0.56]{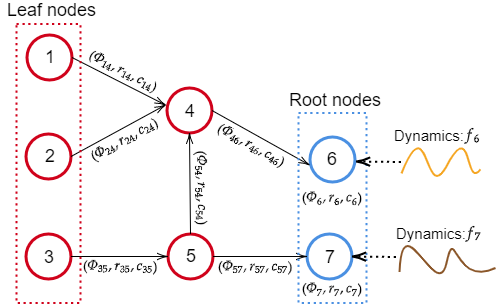}
    \caption{\small{Example of a hybrid attack graph where $\calC=\{1,2,3,4,5\}$ and $\calP=\{6,7\}$ are the set of cyber (red) and physical (blue) nodes.}}\label{fig:hag}
    \vspace{-5mm}
\end{figure}
\subsection{Markov Decision Process Model}\label{sec:mdp}
Here, we describe the main constituents of the MDP model including the state and action spaces, state transitions, reward function, and MDP objective function. 
\subsubsection{State Space}
Let $\calT=\{1,\ldots, T\}$ be a finite attack horizon, where $t\in \calT$ is the $t$-th time period. Let $s_t^i\in\{0,1\}$ be the security state of node $n\in\calN$ at time $t$, where $s_t^i=1$ if the adversary has compromised node $i$ by time $t$, and $s_t^i=0$ otherwise. Let $P$ be the total number of physical nodes. The dynamical state at a physical node $n\in\calP$ at time $t$ is denoted by $x_t^n\in\calX$, while $w_t^n$ is a random disturbance that affects the dynamics $f_n$. Then, the (random) system state at time $t$ is
\begin{equation}\label{eq:state}
    s_t = (s_t^1, \ldots, s_t^N, x_t^1, w_t^1, \ldots, x_t^P, w_t^P)\in\calS,
\end{equation}
where $\calS$ is a bounded, \textit{hybrid} (discrete-continuous) state space. We use $s$ to denote a generic state in $\calS$.
\subsubsection{Action Space}
Define $\calE_t:=\{(i,j)\in\calE: s_t^i=1, s_t^j=0\}$, and $\calP_t:= \{i\in\calP: s_t^i=1\}$, where $\calE_t$ is the set of edges with available cyber exploits and $\calP_t$ is the set of compromised physical nodes at time $t$, respectively. Let $\calB_t=\cup_{e\in\calE_t}a_e$ 
be the set of available cyber exploits at time $t$, where $\calB_t\subseteq\calA_\calE$. The action space at time $t$, denoted by $\calA_t(s_t)$, is the set of all cyber exploits and physical-node actions available at $t$, i.e.,
\begin{equation}\label{eq:actionspace}
\calA_t(s_t) =\{\varnothing\}\cup\calB_t \cup \bigl(\cup_{n\in\calP_t}\calA_n\bigr)\subseteq\calA.
\end{equation}
Note that $\calE_0=\{(i,j)\in\calE: i\in\calC, s_0^i=1\}$ and $\calP_0=\emptyset$. The action at time $t$ is denoted by $a_t$.

\subsubsection{State Transition Function} The attack-success probabilities ($\Phi_t$) and the random disturbances $(w_t^1,\ldots, w_t^P)$ govern the state transitions in the MDP. The dynamical state $x_t^n$ at node  $n\in\calP$ evolves according to the dynamics
\begin{subequations}\label{eq:physicaldynamics}
\begin{align}
    x_{t+1}^n &= f_n(x_t^n, u_t^n, w_t^n, a_t^n, \Phi_t),\\
    u_t^n &= \lambda_n(x_t^n, a_t^n, \Phi_t),
\end{align}
\end{subequations}
\noindent where $u_t^n\in\calU$ is the control input, and $\lambda_n$ is a control law that can be an outcome of an optimization. To describe transitions due to cyber exploits, define the set $\calE_t^n:=\left\{(k,n)\in\calE: s_t^k=1, s_t^n=0\right\}\subseteq\calE_t$ that contains the available edges to reach an uncompromised node $n$ at time $t$; note that $\calE_t^n=\emptyset$ if $s_t^n=1$. Let $\calA_t^n$ be the set of cyber exploits related to the edges in $\calE_t^n$. When $\calE_t^n\neq\emptyset$, the probability that $n$ is compromised, denoted by $p_t^n$, is equal to the probability that at least one available exploit is successful, i.e., $p_t^n=1-\prod_{a\in\calA_t^n}(1-\Phi_t^a)$, where $p_t^n=0$ if $\calE_t^n=\emptyset$. Here, it is assumed that the exploits fail independently of each other. In what follows, the acronym ``w.p." stands for \textit{with probability}. Then, the dynamics for $s_t^n$ is described by
\begin{equation} \label{eq:exploitdynamics}
    s_{t+1}^n = 
    \begin{cases}
    s_t^n,  &\mbox{if\,\,} s_t^n=1, \mbox{\,\,(w.p. 1)},\\
    1,      &\mbox{if\,\,} s_t^n=0, \mbox{\,\,(w.p. $p_t^n$)},\\
    0,      &\mbox{if\,\,} s_t^n=0, \mbox{\,\,(w.p. $1-p_t^n$)},
    \end{cases}
\end{equation}
where the first condition in \eqref{eq:exploitdynamics} follows from the monotonicity assumption. Let $d_t:=(\Phi_t, w_t^n)_{n\in\calP}$. For notational brevity, we jointly express the dynamics in \eqref{eq:physicaldynamics} and \eqref{eq:exploitdynamics} as
\begin{equation}\label{eq:dynamics}
    s_{t+1} = g(s_t, a_t, d_t),
\end{equation}
where $g$ is a probability transition kernel.

\subsubsection{Reward Function} Let $r_t(s,a)$ be the random net-reward (reward minus cost) at time $t$ for state $s$ and action $a$. Then,
\begin{equation}\label{eq:reward}
    r_t(s,a)=
    \begin{cases}
    r^a - c^a,              &\mbox{(w.p. $\Phi_t^a$)},\\
    r^\varnothing - c^a,    &\mbox{(w.p. $1-\Phi_t^a$)},
    \end{cases}
\end{equation}
where $r^a=r_e^a$ and $c^a=c_e^a$ if $a$ is a cyber exploit, i.e., $a\in\calB_t$. When $a$ is a physical-node action, i.e., $a\in \calA_n$ for $n\in\calP_t$, then $r^a=r_n^a$ and $c^a=c_n^a$ in \eqref{eq:reward}. The rewards $r_e^a, r_n^a$ and costs $c_e^a, c_n^a$ were defined in Definition \ref{def:hag}.

\subsubsection{Objective Function}
\noindent Let $\pi_t:\calS \to \Delta_{\calA_t}$ denote a policy that maps a state $s$ to a probability distribution over the action space $\calA_t(s)$. Note that this also includes the space of deterministic policies $\pi_t^d:\calS \to \calA_t$. We seek an attack policy of the form $\pi=(\pi_t)_{t\in\calT}$, such that $a=\pi_t(s)\in\calA_t(s)$. Let $\Pi$ be the space of all feasible policies. Starting from an initial state $s$, the adversary seeks a policy $\pi^*\in\Pi$ that maximizes the total expected finite-horizon reward, i.e.,
\begin{equation}\label{eq:mdpobjective}
    \pi^*\in\argmax_{\pi\in\Pi}\,\E\left[\sum_{t\in\calT}r_t(s_t, \pi_t(s_t))\Big{|}\pi, s_0=s\right],
\end{equation}

\noindent where the expectation is taken with respect to the transition kernel in \eqref{eq:dynamics}. Using the linearity of the expectation operator, Equation \eqref{eq:mdpobjective} can be written in the following recursive form, known as the Bellman optimality equation~\cite{sutton2018reinforcement}:
\begin{equation}\label{eqn:bellmans}
    V_t^*(s) = \max_{a\in\calA_t(s)}\E(r_t(s,a) + V_{t+1}^*(s_{t+1}|s,a)),
\end{equation}
\noindent where $V_t^*$ is the optimal value function at time $t$. The optimal policy $\pi^*$ can be extracted from $V_t^*$ using
\begin{equation}
\pi_t^*(s) \in \argmax_{a\in\calA_t(s)}\,\E(r_t(s,a) + V_{t+1}^*(s_{t+1}|s,a)).   
\end{equation}
\noindent Note that $V_t^*(s)$ denotes the expected total return when, starting from state $s$ at time $t$, an adversary follows $\pi^*$ till the end of the horizon.
\subsection{Computational Challenges}\label{sec:computational}
\noindent We now emphasize the major challenges in solving the MDP in \eqref{eq:mdpobjective}. First, note that classical dynamic programming (DP) algorithms, such as value- and policy-iteration~\cite{sutton2018reinforcement}, are not amenable for solving \eqref{eq:mdpobjective} as they require multiple sweeps over the state space to solve the optimality equations in \eqref{eqn:bellmans}, which is clearly intractable as $\calS$ is uncountable. Second, DP methods assume perfect knowledge of the transition model in \eqref{eq:dynamics}. However, an adversary usually has limited knowledge of the dynamics in \eqref{eq:physicaldynamics} and the attack success probabilities. Next, we discuss two competing reinforcement learning (RL) algorithms and one greedy baseline procedure to approximately solve~\eqref{eq:mdpobjective}.


\section{Solution Approaches}\label{sec:policies}
\subsection{Approximate Dynamic Programming}\label{sec:adp}

Approximate dynamic programming (ADP) is a model-based RL approach that overcomes the challenge of a hybrid MDP state space.  Unlike DP methods that use backward enumeration, ADP steps forward in time and generates sample state trajectories to learn a low-dimensional parametric approximation of $V_t^*$.  Let $\theta\in\Theta$ be a parameter vector, where $\Theta$ has a significantly smaller dimension than that of $\calS$. The goal in ADP is to iteratively learn a value $\theta^*\in\Theta$ such that $\forall s \in \calS, |V_t^*(s)-J_t(s;\theta^*)|<\epsilon$, where $J_t(s;\theta)$ is a parametric function of $\theta$, and $\epsilon>0$ is a small tolerance value. Let $k$ be the iteration index, $s_t^k$ denote the sampled state at time $t$ in iteration $k$, and $\hat\theta$ be the current estimate of $\theta^*$ at start of iteration $k$. Assuming an attacker has knowledge of the dynamics in \eqref{eq:dynamics}, a sample estimate of $V_t^*$ at state $s_t=s_t^k$ is 
\begin{equation}\label{eq:adpvalueestimate}
    \hat v^k_t = \max_{a\in\calA_t(s_t^k)} \E\left[r_t(s_t^k,a) + J_{t+1}(s_{t+1}^k;\hat\theta)\right].
\end{equation}
Note that \eqref{eq:adpvalueestimate} evaluates a greedy action with respect to (w.r.t.) the current function approximation evaluated at the sampled state, without requiring a complete sweep of the state space. Once $\hat v_t^k$ is computed, $\theta$ is immediately updated using an online stochastic gradient algorithm as follows: 
\begin{equation}\label{eq:SGD_ADP}
    \hat\theta\leftarrow\hat\theta+ \alpha\nabla_{\theta}\left(J_t(s_t^k;\hat\theta) - \hat v_t^k\right),
\end{equation}
where $\alpha>0$ is a step size that varies over the iterations, and $\nabla_\theta$ is the gradient of $J_t$ w.r.t. $\theta$ evaluated at $(s_t^k,\hat\theta)$. The process is repeated until $\hat\theta$ converges or a prescribed number of iterations is completed. The update in \eqref{eq:SGD_ADP} is guaranteed to converge if $\sum_{i=1}^\infty\alpha_i=\infty$ and $\sum_{i=1}^\infty\alpha_i^2<\infty$~\cite{powell2009you}. The steps of the ADP algorithm are described in the Appendix. Although ADP is applicable even if the dynamics are unknown or mis-specified, we use the model-based version here to compare it against pure model-free RL methods, such as the Actor-Critic algorithm that is discussed next.
\subsection{Actor-Critic Algorithm} \label{sec:a2c}
The Actor-Critic (AC) algorithm~\cite{sutton2018reinforcement} is an iterative model-free procedure that concurrently trains two models (called the actor and the critic) to learn a parametric form of the optimal policy of \eqref{eq:mdpobjective}, without requiring any knowledge of the transition dynamics in \eqref{eq:dynamics}. Let $\pi_t(a|s;\psi)$ denote a stochastic policy at time $t$ parameterized by $\psi\in\Psi$, and let $J_t(s;\theta)$ be the corresponding value function approximation, as defined in Section~\ref{sec:adp}. At each time step of a given episode, the \textit{critic} updates the value-function parameters $\theta$ using sampled actions and successor states, while the \textit{actor} updates the policy parameters $\psi$ in a direction suggested by the critic. A stochastic gradient scheme (similar to~\eqref{eq:SGD_ADP}) updates both $\psi$ and $\theta$. The process is repeated for different episodes and terminates once a prescribed convergence criterion is met.


The AC algorithm is most suited for problems with continuous action spaces. To apply it for the MDP in \eqref{eq:mdpobjective} with discrete actions, an exponential softmax distribution is used as the parametric form for $\pi_t$, i.e., for each $t\in\calT$,
\begin{equation}\label{eq:soft-max-def}
    \pi_t(a|s;\psi) =\dfrac{e^{h(s,a,\psi)}}{\sum_{b\in\calA_t(s)} e^{h(s,b,\psi)}}, \quad \forall a \in \calA_t(s),
\end{equation}
where $e$ is the Euler constant. The function $h(s,a,\psi)$ in \eqref{eq:soft-max-def} denotes a real-valued parametric preference defined for each state-action pair, which can be encoded using tile coding or deep neural networks. The main steps of the AC algorithm are described in the Appendix.


\subsection{Greedy Attack Policy}\label{sec:greedypolicy}
We discuss a greedy attack scheme for hybrid attack graphs that serves as a baseline for the aforementioned RL procedures. The proposed greedy policy exploits the topology of a HAG to identify a reduced set of available actions, which is then used to  execute cross-layer attacks in a myopic fashion. Next, we describe the main steps of the greedy policy in more detail.



For a given cyber exploit $a$, let $R_a$ be the set of successive actions an adversary can use to reach any root node in $\calP$. For example, in Figure~\ref{fig:hag},  the action $a=a_{3,5}$ along the edge $(3,5)$ has an associated $R_a=\{a_{3,5}, a_{5,4}, a_{4,6}, a_{5,7}\}$.  Starting from an initial state $s$ at time $t=0$, the value in using a cyber exploit $a\in\calB_0$ to reach any of the root nodes is expressed as
\begin{equation}\label{eq:pathcondexpvalue}
    Q(a,s) = \sum_{\tilde a\in R_a} \E[r_0(s,\tilde a)],
\end{equation}
where the expectation is taken w.r.t to the success probabilities in~\eqref{eq:reward} but not on the security states. We extend the definition in \eqref{eq:pathcondexpvalue} to a set of 
cyber exploits $A\subseteq\calA_\calE$ as 
\begin{equation}\label{eq:pathtotalvalue}
    Q(A, s) = \sum_{a\in R_A} \E[r_0(s, a)].
\end{equation}

\noindent Starting with an empty set $A$, the greedy policy iteratively adds the most valuable cyber exploits to $A$ by maximizing $Q(A,s)$ in \eqref{eq:pathtotalvalue}, i.e.,
\begin{equation}
   a^*=\argmax_{a\in\calB_0\setminus A}\,\, Q(A\cup\{a\}),~
   A \leftarrow A\cup \{a^*\}.\label{eq:greedypathselection}
\end{equation}

\noindent The number of actions to be added to $A$, denoted by $\ell$, is set apriori. The final output $A$ provides an adversary with the $\ell$ most valuable (initial) exploits to reach the root nodes. For example, if $\ell=1$, the adversary selects an available exploit that yields the largest reward along a path to a root node. Starting from an initial state $s_0$ and using the initial exploits in $A$, the adversary now executes a myopic policy at each time $t$, denoted by $\pi_t^g$, that maximizes the nominal net-reward, i.e.,
\begin{equation}\label{eqn:greedyonesteppolicy}
    \pi_t^g(s_t) \in \argmax_{a\in\calA_{t}(s_t)\cap A} \{r^a - c^a\},
\end{equation}
where $\calA_{t}(s_t)\cap A$ is the reduced action space at $t$. The steps of the Greedy attack policy is provided in the following table.

\begin{algorithm}
\SetAlgoLined
\textbf{Phase 1: Pruning initial cyber exploits}\\
\textbf{Input}: $\calG, \ell, s_0$ \\
\KwResult{Set of valuable cyber exploits $A$}
 Initialize $A\leftarrow\textnormal{empty}$;\\
 \While{$|A| < \ell$:}{
  select an exploit $a^*$ and update $A$ using \eqref{eq:greedypathselection};
  }
\textbf{Phase 2: Myopic policy execution}\\
\textbf{Input:} $A, s_0$ \\
\For{$t \in \calT$:}{
    select action $a_{t} = \pi_{t}^{g}(s_{t})$ using \eqref{eqn:greedyonesteppolicy};\\  transition to the next state according to \eqref{eq:dynamics};\\
    }
 \caption{Greedy Attack Policy}\label{alg:greedyattack}
\end{algorithm}

\section{Results and Discussion}\label{sec:results}
\subsection{Use Case: Sensor Deception Attacks in Buildings}
\begin{figure}[htb!]
    \vspace{-5mm}
    \centering
    \includegraphics[scale=0.37]{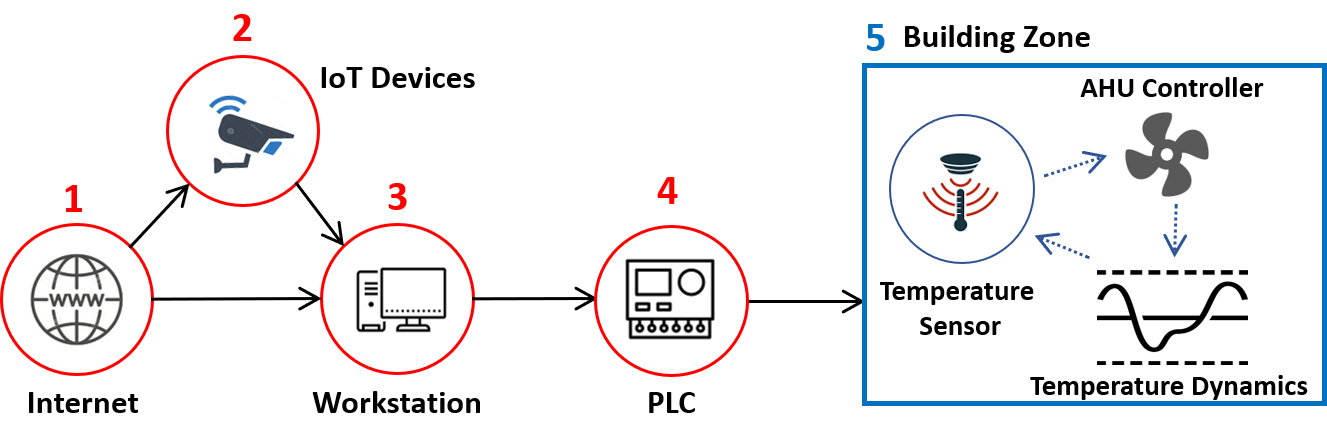}
    \caption{\small{The proposed HAG for building sensor deception attacks. The red circles depict the cyber nodes, while the blue rectangle is the root node representing the building zone under attack.}}
    \label{fig:building_hag}
\end{figure}

We consider sensor-deception attacks in buildings where an adversary targets to maximize occupant discomfort. Regular building operations involve an air-handling unit (AHU) that re-conditions ambient air to a specific supply-air temperature, which is then forced into the building zones by a supply fan. The adversary seeks an intelligent way to manipulate temperature measurements (from zone-level sensors) to deceive the AHU control in sending
poorly conditioned air into the zones, causing large comfort-bound violations over time. However, to access the temperature sensors, the adversary has to first execute a set of cyber exploits on different components of a Building Automation System (BAS), including IoT devices (e.g. IP camera and smart thermostats), building-management workstations, and programmable logic controllers (PLC). For demonstrative purposes, we assume that only a single zone, with a dedicated AHU and temperature sensor, is under attack. For our use case, we use the HAG in Figure \ref{fig:building_hag} as a proof-of-concept. Similar attack graphs for BAS were used in~\cite{dosSantos2019}.

\noindent Next, we describe the cyber and physical layer components of the proposed HAG using notation similar to Section \ref{sec:modelformulation}.
\subsubsection{Cyber Layer} The HAG consists of four cyber nodes and five cyber exploits. Table~\ref{table:cyber_layer} tabulates the different cyber exploits and associated success probabilities (assumed to be time-invariant). Each exploit incurs a cost ($c_e^a$) of 0.1. A nominal reward ($r_e^a$) of 1 is gained if an exploit is successful, while the reward for doing nothing ($r_e^\varnothing$) is set to  0. 

\begin{table}[htb!]
\centering
\caption{\small{Description of the cyber exploits and associated success probabilities in the cyber layer.}}
\label{table:cyber_layer}
\begin{tabular}{|c|c|c|c|} 
\hline
\textbf{Edge} & \textbf{Type of Cyber Exploit} & \textbf{Success Probability} \\ 
\hline
(1, 2) &Initial access to node 2    &0.9\\
\hline
(1, 3) &Initial access to node 3    &0.7\\ 
\hline
(2, 3) &Lateral movement to node 3  &0.9\\
\hline
(3, 4) &Lateral movement to node 4  &0.8\\
\hline
(4, 5) &Command and control from node 4 &0.5\\
\hline
\end{tabular}
\end{table}

\subsubsection{Physical Layer}
 Let $x_t\in\R$ be the zone temperature (in $^{\circ}$C) at time $t$ and $u_t=(z_t, m_t)$ denote the vector of AHU control inputs, where $z_t\in\R$ is the supply-air temperature (in $^{\circ}$C) and $m_t\in\R_+$ is the airflow rate (in kg/s) at time $t$, respectively. The (random) outside-air temperature (in $^{\circ}$C) at time $t$ is denoted by $w_t\in\R$. Similar to~\cite{dong2019online}, the zone temperatures evolve according to the nonlinear dynamics
\begin{equation}\label{eq:zonedynamics}
    x_{t+1} = x_t + 0.01 m_t(z_t-x_t) + 0.1 (w_t-x_t).
\end{equation}

\noindent Let $x^L=23^\circ$C and $x^U=25^\circ$C be the lower and upper thermal-comfort bounds, respectively, where an occupant is comfortable if $x_t\in[x^L, x^U]$, and is uncomfortable otherwise. The outside-air temperature at each time step is sampled from
\begin{align}\label{eqn:toa}
    w^k_t \sim 0.5(x^L + x^U) + 4\sin(0.125t + k) + \phi_t,
\end{align}

\noindent where $\phi_t \sim U(-1,1)$ is a uniform random variable between [-1,1] and $k$ denotes the phase shift of the sine wave. With a slight abuse of notation, let $a_t\in\calA_5$ denote the adversarial perturbation (in $^\circ$C)  at time $t$. We assume the success probabilities to be independent of the attack actions and decreases monotonically over time 
\begin{equation}\label{eq:attackprob}
    \Phi_t = 0.5 - 0.1\lfloor t/10\rfloor,
\end{equation}
where $\lfloor\cdot\rfloor$ is the floor function. Let $b_t\sim B(\Phi_t)$ be a Bernoulli random variable with parameter $\Phi_t$. The perturbed measurement at time $t$ equals $y_t=x_t+a_tb_t$, where $y_t=x_t+a_t$ if $a_t$ is successful (w.p. $\Phi_t)$, and $y_t=x_t$ otherwise (w.p. $1-\Phi_t$). The control variables in $u_t$ are set according a threshold policy that depends on $y_t$ as follows: 
\begin{equation}\label{eq:thresholdpolicy}
    (z_t, m_t)=
    \begin{cases}
    (30, 10\min\{x^L-y_t,1\}),    &\mbox{if $y_t<x^L$},\\
    (15, 10\min\{y_t-x^U,1\} )    &\mbox{if $y_t>x^U$},\\
    (z_{t-1}, 0)                  &\mbox{otherwise}.
    \end{cases}
\end{equation}
\noindent Using \eqref{eq:thresholdpolicy}, the dynamics in \eqref{eq:zonedynamics} can be expressed as $x_{t+1}=f(x_t,w_t,a_t)$. Define $g(u)\equiv\max\{u,0\}$ for $u\in\R$. Then, the reward for executing action $a_t$, denoted by $r(a_t)$, equals
\begin{equation*}
    r(a_t)=g\left(x^L-f(x_t,w_t,a_t)\right)+g\left(f(x_t,w_t,a_t)-x^U\right),
\end{equation*}
where the first (resp. second) term  is the thermal discomfort caused by temperature deviation from the lower (resp. upper) comfort bound. The cost of executing an action $a_t$ is set to $0.5a_t^2$. Therefore, the net-reward at time $t$ equals
\begin{equation}\label{eq:rewardbuildings}
    r_t(s_t,a_t)=
    \begin{cases}
    r(a_t) - 0.5a_t^2,               &\mbox{(w.p. $\Phi_t$)},\\
    -0.5a_t^2,                    &\mbox{(w.p. $1-\Phi_t$)}.
    \end{cases}
\end{equation}
\subsection{Experimental Setup}
We used tile coding~\cite{sutton2018reinforcement} to construct a sparse feature representation of the state space. The value functions were defined as linear-function approximations over the set of tiles. The number of time steps in the attack horizon was set to 48. The comfort range $[x^L, x^U]$  was kept fixed over the entire horizon. The attack policies and the corresponding value functions for the ADP and the AC procedures were trained over 50,000 episodes. For each episode $k$, sample trajectories of the outside-air temperature and success probabilities were generated using \eqref{eqn:toa} and \eqref{eq:attackprob}, respectively. Note that the phase shift in \eqref{eqn:toa} is incremented over the episodes to avoid over-fitting to a specific outside temperature trajectory. The initial zone temperatures were sampled uniformly from the range $[x^L,x^U]$. Once the training was complete, the performance of all three policies was compared over 10,000 test episodes. 



\subsection{Results and Discussion}
Table \ref{table:attack_time} compares the average number of time steps (over the test episodes) that it took to access the physical node under the ADP, AC, and Greedy policies. Due to their predictive capabilities, the ADP and AC policies secured quicker access to the physical node to cause occupant discomfort early in the attack horizon. By contrast, the Greedy policy prioritizes short-term gains accrued by executing all of the cyber exploits, which delays the corresponding access to the physical node.


\begin{table}[htb!]
\centering
\caption{\small{Average number of time steps to reach the physical node.}}
\label{table:attack_time}

\begin{tabular}{|M{2.3cm}|M{4cm}|} 
\hline
\textbf{Attack policy} & \textbf{Avg. time to reach root node}\\ 
\hline
Greedy &10.7\\
\hline
ADP &8.4\\
\hline
AC & 8.7\\
\hline
\end{tabular}
\end{table}

\begin{figure*}[htb]
	\centering
		\subfloat[ADP Policy]{\includegraphics[scale=0.32]{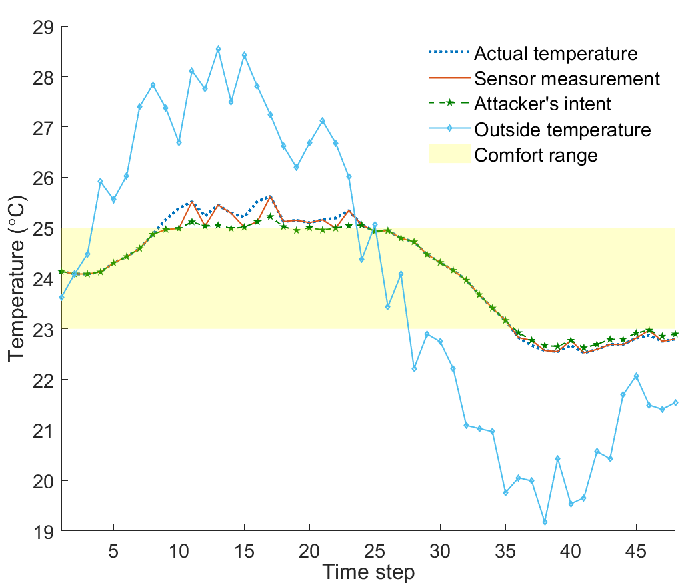}
			\label{fig:adp_sample}}
		\subfloat[Actor-Critic Policy]{\includegraphics[scale=0.32]{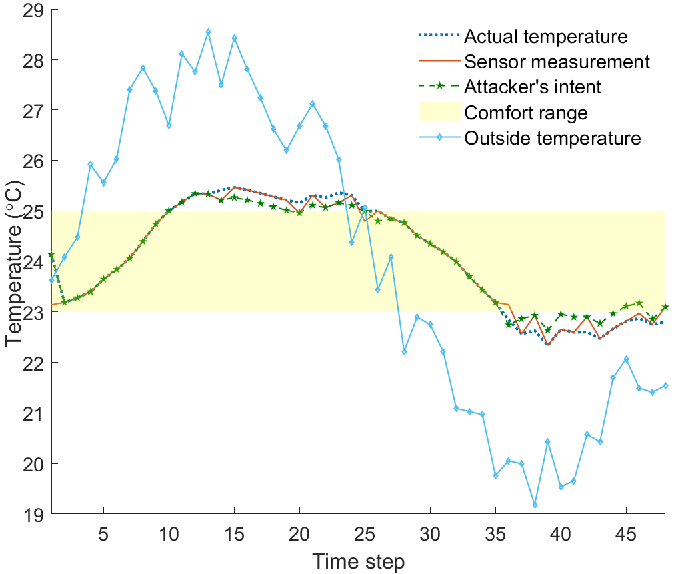}
			\label{fig:actor_sample}}
		\subfloat[Greedy Policy]{\includegraphics[scale=0.32]{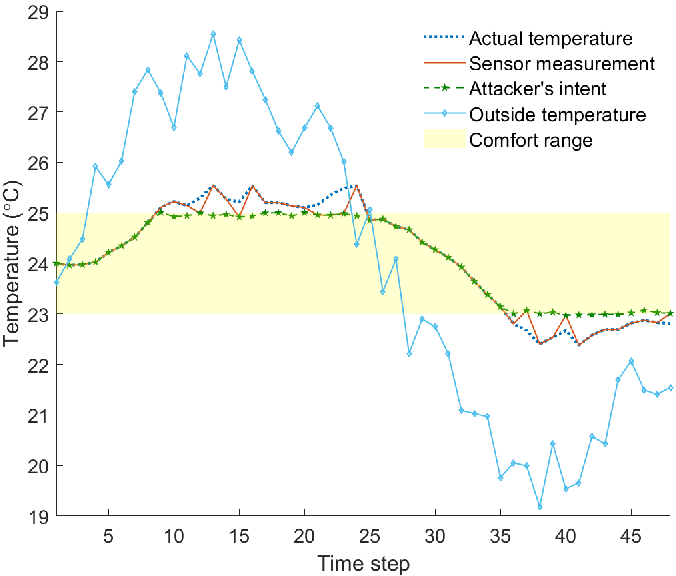}
			\label{fig:greedy_sample}}
		\caption{Illustration of sample trajectories using the learned policies from ADP, AC, and Greedy algorithms.}\label{fig:timeseries}
	\vspace{-0.2in}
\end{figure*}

\noindent Figure \ref{fig:timeseries} depicts the performance of the three attack policies at the physical node for a representative test trajectory. Note at each time $t$, the sensor measurement ($y_t$) is bounded by the actual zone temperature ($x_t$) and the attacker's intent ($x_t+a_t$), depending on whether action $a_t$ is successful or not. In general, all three policies seeks to deceive the AHU controller by maintaining the sensor measurements near the comfort bounds and keeping the true temperatures outside the comfort bounds over longer durations. As the success probabilities decay towards the end of the horizon, the greedy policy prescribes high-risk, high-reward actions that result in lower average performance towards the end. This is evident in Fig \ref{fig:greedy_sample} where there is a consistent mismatch between the attacker's intent and the sensor measurement towards the end of the horizon. By contrast, both the ADP and AC policies reduce the frequency and the magnitude of attacks towards the end of the horizon where the success probabilities are low (see Figures \ref{fig:adp_sample} and \ref{fig:actor_sample}). We compare the performance of the algorithms using a metric $\rho$ that measures the attacker's overall return in investment, defined as
\begin{equation*}
   \rho = \dfrac{\sum_{t\in\calT} \hat{r}(a_t)}{\sum_{t\in\calT} 0.5 a_t^2}, 
\end{equation*}
where $\hat{r}(a_t)$ is the reward observed after taking an action $a_t$. For the trajectories shown in Fig \ref{fig:timeseries}, the computed return is $13.84$ (ADP), $11.3$ (AC), and $4.03$ (Greedy). 

To study the impact of the size of action space on policy performance, we assume $a_t \in \{-2,-2+\delta,\ldots, 2-\delta, 2\}$, where different values of $\delta \in \{1/s\,|\, s\in \{1, 3,\ldots,9\}\}$ produce different action spaces. Table \ref{table:sweep_data} highlights the performance of the different algorithms for different sizes of the action space. Note that action space becomes larger, both the ADP and the AC policies exhibit similar performance, while outperforming the greedy algorithm. 

\begin{table}[htb!]
\centering
\caption{\small{The mean and variance of the net reward for different action-space dimensions}}
\label{table:sweep_data}
\begin{tabular}{|c|c|c|c|c|c|c|} 
\hline
\multirow{2}{*}{\textbf{Number of}\vspace{0.2cm}} & \multicolumn{3}{|c|}{\textbf{Mean of net reward}} & \multicolumn{3}{|c|}{\textbf{Variance of net reward}} \\
\cline{2-7}
\textbf{actions}&ADP&AC&Greedy&ADP&AC&Greedy\\ \hline
5&7.2&6.6&6.09&0.49&0.75&0.804 \\ \hline
13&7.8&7.9&6.0&0.882&1.61&1.51 \\ \hline
21&8.03&7.9&6.3&0.861&1.56&1.85 \\ \hline
29&8.06&8&6.37&0.876&1.49&1.99 \\ \hline
37&8.09&7.96&6.39&0.881&1.84&2.04 \\ \hline
\end{tabular}
\end{table}
\section{Conclusion and Future Work}\label{sec:conclusion}
In this paper, an MDP-based approach is adopted to determine the optimal attack sequence over a hybrid attack graph having both discrete and continuous components. Two RL algorithms were implemented to overcome the computational challenges of a hybrid state MDP. A greedy attack scheme was developed as a baseline for the RL procedures. Finally, a building use-case is used to demonstrate the comparative performance of the proposed algorithms. 

In the future, this methodology will be extended to cases in which an attacker may have limited information of the attack graph using partially observable MDP models. Hybrid model-based and model-free RL algorithms will be developed to guide the learning process in cases where the CPS dynamics are complex and accurate system models are unavailable. Finally, deep RL methods will be explored to solve larger hybrid attack graphs of complex CPS.

\section*{Acknowledgements}
\noindent This research is part of a sponsored project under the  Mathematics for Artificial Reasoning in Science (MARS) initiative at the Pacific Northwest National Laboratory, USA.
\bibliographystyle{IEEEtran}
\bibliography{refs}

\appendix
\noindent The appendix is available at: https://bit.ly/3odSMce

\end{document}


\title{Appendix to ``Automated Adversary Emulation for Cyber-Physical Systems via Reinforcement Learning"}
\author{
\IEEEauthorblockN{
Arnab Bhattacharya\IEEEauthorrefmark{1}, 
Thiagarajan Ramachandran\IEEEauthorrefmark{1}, 
Sandeep Banik\IEEEauthorrefmark{3}, 
Chase P. Dowling\IEEEauthorrefmark{2}, 
Shaunak D. Bopardikar\IEEEauthorrefmark{3}}
\IEEEauthorblockA{
\IEEEauthorrefmark{1}\textit{Optimization and Control Group}, \IEEEauthorrefmark{2}\textit{Information Modeling and Analytics Group} \\ 
\textit{Pacific Northwest National Laboratory}, Richland, WA, USA}
\IEEEauthorblockA{\IEEEauthorrefmark{3}\textit{Electrical and Computer Engineering}, 
\textit{Michigan State University}, East Lansing, MI, USA}
}
\maketitle
\begin{algorithm}[h]
	\textbf{Input:} Parameterized state-value function $J_t(s_t, \theta)$ \\
	\textbf{Parameters:} Step-size $\epsilon$, discount factor $\gamma$, learning rate $\alpha$\\
	\textbf{Initialize:} state-value weights $\theta \in \mathbb{R}^{d} = 0$,  \\
	\For{ every episode $k$}{
	Select a sample path $d^k$ for all uncertainties \\
	Initialize $s_0 \in [s_{min}, s_{max}]$\\
	    \For{$t \gets 0$ to $T$}{
	        $\hat v^k_t \leftarrow \max_{a\in\calA_t(s_t^k)} \E\left[r_t(s_t^k,a) + \gamma J_{t+1}(s_{t+1}^k;\theta)\right]$\\
	        Let $a_t^k \in \mathcal{A}_t(s_t^k)$ be the corresponding optimal action \\
	        $v^k_t \leftarrow \alpha v^k_t + (1-\alpha) J_t(s_t^k, \theta)$ \\
	        \theta\leftarrow\theta+ \epsilon\nabla_{\theta}\left(J_t(s_t^k;\theta) - \hat v_t^k\right)\\
	        $s^k_{t+1} \leftarrow g(s^k_t, a^k_t, d^k_t)$ \\
	    }
	}
	\caption{Approximate dynamic programming method}
	\label{algo:ADP}
\end{algorithm}

\begin{algorithm}[h]
	\textbf{Input:} Parameterized policy $\pi(a|s_t,\psi)$,~(\ref{eq:soft-max-def}) \\
	\textbf{Input:} Parameterized state-value function $J(s_t,\theta)$, \\
	\textbf{Parameters:} Step-size $\alpha^{\theta} > \alpha^{\psi} > 0$, discount factor $\gamma$  \\
	\textbf{Initialize:} Policy parameter $\psi \in \mathbb{R}^{d'} = 0$, state-value weights $\theta \in \mathbb{R}^{d} = 0$,  \\
	\For{ every episode}{
	Initialize $s_0 \in [s_{min}, s_{max}], I \leftarrow 1$ \\
	    \For{$t \gets 0$ to $T$}{
	        $a_t \sim \pi(.|s_{t}, \psi)$ \\
	        $s_{t+1} \leftarrow g(s_t, a_t, d_t)$~ \\
	        $r_{t+1} \leftarrow r(s_t, a_t)$~\\
	        $\delta \leftarrow r_{t+1} + \gamma J(s_{t+1},\theta) - J(s_{t},\theta)$ \\
	        $\theta \leftarrow \theta + \alpha^{\theta}\delta\gamma\nabla J(s_{t},\theta)$ \\
	        $\psi \leftarrow \psi + \alpha^{\psi}I\delta\gamma\nabla \text{ln} \pi(a_t|s_t,\psi)$ \\
	        $I \leftarrow \gamma I$
	        
	    }
	}
	\caption{One-step Actor-Critic method}
	\label{algo:A2C}
\end{algorithm}

\section{Greedy attack policy performance guarantees}

Here we describe how can leverage the performance guarantees of greedy path selection proved in \cite{saha2016identifying}. Recall the output set of actions $A$ of Algorithm~Greedy Attack Policy applied to an attack graph $\calG$ yields a subgraph $\calG'$ induced by all root physical nodes $\calP$ reachable via the set of actions $R_{a}$, reduces the number of actions available to a strict application of the myopic policy $\pi^{g}$.

\begin{lemma} [\cite{saha2016identifying}, Lemma III.2 and III.3] If a given attack graph $\calG$ with time horizon $\calT$ is 1) directed acyclic and 2) composed only of OR preconditions, then Algorithm Greedy Attack Policy applied at each $t\in\calT$, yields the following bound on the expected pathwise rewards,
\[
\sum_{t \in \calT} Q(a'_{t}, s_{t}) \geq \big (1 - \frac{1}{e} \big) \sum_{t \in \calT} Q(a^{*}_{t}, s_{t}).
\]
\end{lemma}
\begin{proof}
Given a HAG $\calG$, fix $t=1$. By Lemma III.3 of [Saha 2016], algorithm Greedy Attack Policy applied to $\calG$ at time $t$ yields a subgraph $\calG'$ such that $a_{t} \in \calA_{t}'(s_{t})$, the action set of subgraph $\calG'$. We have then an action chosen according to the MDP solution in $\calG'$, $a_{t}'$, and corresponding the optimal action over the original HAG $\calG'$, $a_{t}^{*}$. Consequently, $Q(a'_{t}, s_{t}) \geq (1 - \frac{1}{e})Q(a^{*}_{t}, s_{t})$. By repeated application of Alg.~Greedy Attack Policy at each $t \in \calT$, a sequence $z = \{a'_{t}, s_{t}\}_{t = 1}^{\calT}$ is realized according to MDP solution method, and

\begin{equation}
\sum_{t \in \calT} Q(a_{t}', s_{t}) \geq \left(1 - \frac{1}{e}\right) \sum_{t \in \calT} Q(a_{t}^{*}, s_{t})
\end{equation}

\end{proof}

In words, greedily pruning the attack graph at each time-step $t$ (or any subset of $\calT$, e.g. the singleton set $t=1$) according to the sum of expected instantaneous rewards conditioned on the current state $s_{t}$, any action choice over the reachable paths $R_{a_{t}}$ for $a_{t} \in \calA$ is at least $(1 - \frac{1}{e})$ of the true optimal solution at time $t$.

\bibliographystyle{IEEEtran}
\bibliography{refs}